\pdfoutput=1

\documentclass[11pt]{article}

\usepackage[final]{acl}

\usepackage{times}
\usepackage{latexsym}

\usepackage[T1]{fontenc}

\usepackage[utf8]{inputenc}

\usepackage{microtype}

\usepackage{inconsolata}

\usepackage{graphicx}
\usepackage{booktabs}
\usepackage[export]{adjustbox}
\usepackage{multirow}
\usepackage{url}
\usepackage{tabularx}
\usepackage{ulem} 
\usepackage{enumitem} 
\usepackage{amsmath}

\title{Process-Supervised Reward Models for Verifying Clinical Note Generation: A Scalable Approach Guided by Domain Expertise}

\author{
Hanyin Wang$^{1,2}$, Chufan Gao$^2$, Qiping Xu$^1$, Bolun Liu$^1$, Guleid Hussein$^1$,  \\ \textbf{Hariprasad Korsapati$^1$, Mohamad El Labban$^1$, Kingsley Iheasirim$^1$}, \\ \textbf{Mohamed Hassan$^1$, Gokhan Anil$^1$, Brian Bartlett$^1$, Jimeng Sun$^{2,3}$}\\
\\
$^1$ Mayo Clinic Health System, $^2$ School of Computing and Data Science, UIUC \\ $^3$ Carle Illinois College of Medicine, UIUC\\
\texttt{wang.hanyin@mayo.edu, jimeng@illinois.edu}
}

\begin{document}

\maketitle

\begin{abstract}
Process-supervised reward models (PRMs) excel at providing step-by-step verification for large language model (LLM) outputs in domains like mathematics and coding. However, their application to fields lacking ground-truth answers, such as clinical note generation, poses significant challenges. We introduce a novel framework for training PRMs to deliver step-level reward signals for LLM-generated clinical notes. By precisely defining meaningful ``\textbf{steps},'' injecting realistic ``\textbf{errors}'' informed by domain expertise, and leveraging LLMs to generate process supervision data at scale, we overcome previous limitations. \textbf{Our PRM, built on LLaMA-3.1 8B, consistently outperforms proprietary reasoning and non-reasoning models}, achieving state-of-the-art (SOTA) performance on two key evaluations: (1) distinguishing gold-standard from error-containing samples with 98.8\% accuracy, and (2) selecting physician-preferred clinical notes with 56.2\% accuracy. We investigate critical components for effective PRM training, including optimal loss functions and data selection strategies, and present a comprehensive physician reader study identifying predictors of downstream Best-of-N performance. Our study sheds light on unlocking the potential of PRMs for diverse generative tasks across domains.\footnote{Our code is available at \url{https://github.com/hanyin88/prm-clinic}.}

\end{abstract}

\section{Introduction}\label{introduction}

\begin{figure*}[!ht]
    \captionsetup{labelfont=bf={bf},skip=12pt}
    \centering
    \includegraphics[width=1\linewidth]{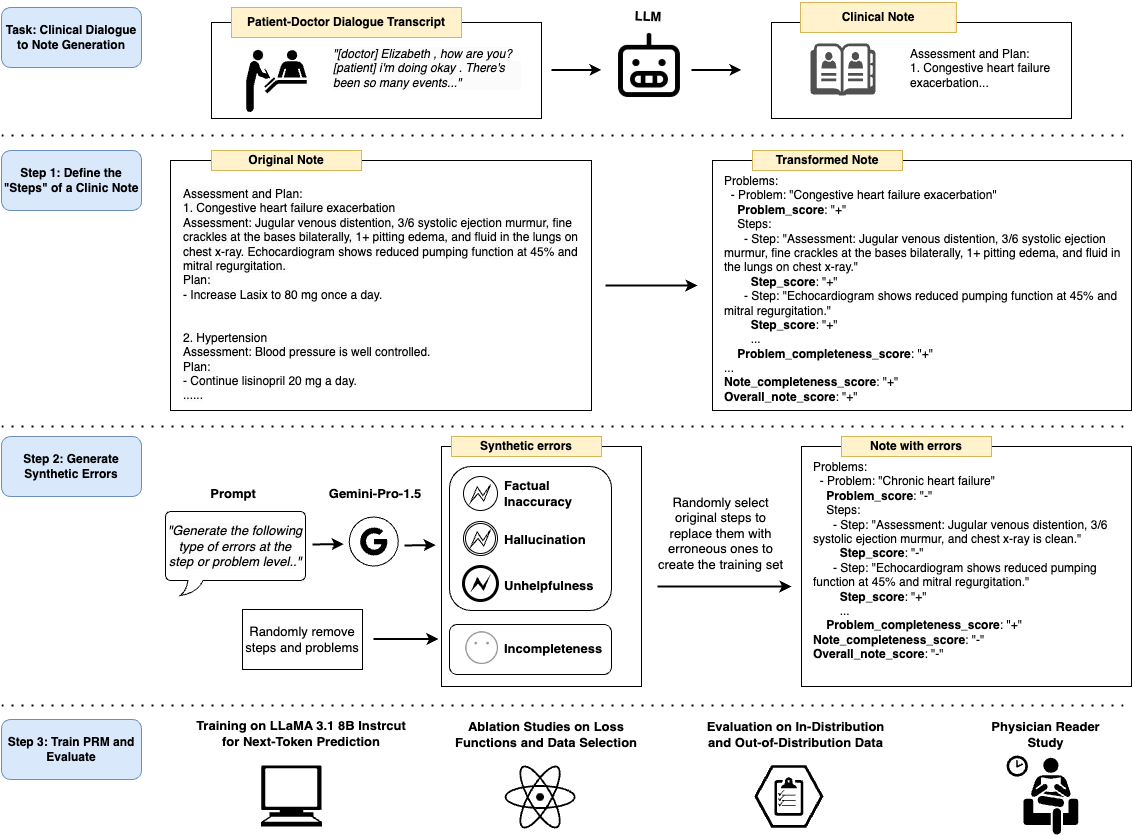}
    \caption{\textbf{Overview of study design.} We train a PRM to step-by-step verify clinical notes generated by LLMs from patient-doctor dialogues. \textbf{Step 1}: Clinical notes are transformed into a hierarchical structure of steps, designed based on domain expertise to capture the key considerations in clinical documentation. \textbf{Step 2}: Gemini Pro 1.5 is utilized to generate synthetic errors from predefined categories. These errors are systematically swapped with the original steps to create negative samples. \textbf{Step 3}: The PRM is trained using LLaMA-3.1 8B instruct, followed by ablation studies and a physician reader study.}
    \label{fig:overview}
    \vspace{-10pt}
\end{figure*}

LLMs show promise in generating clinical notes~\cite{van2024adapted}, but their outputs may contain errors and misalign with physician preferences~\cite{omiye2024large, jin2024hidden}. Currently, no automated, scalable method exists to evaluate the quality of LLM-generated clinical notes, leaving manual evaluation the gold standard. Consequently, costly clinician reader studies are required to validate LLM-based products, such as ambient AI scribes~\cite{liu2024does}. This challenge is amplified by the rapid adoption of ambient scribing technology in the real world~\cite{tierney2025ambient}, with projections indicating that 30\% of the healthcare market will use LLM-generated clinical notes by 2025~\cite{beavins2024ai}.

A potential solution to this challenge is to verify LLM outputs using a reward model (RM). RMs are integral to reinforcement learning for LLM post-training \cite{ouyang2022training, stiennon2020learning}. When employed as verifiers, RMs are categorized into two types: outcome-supervised reward models (ORMs) \cite{cobbe2021training} and, more recently, process-supervised reward models (PRMs) \cite{lightman2023let}. ORMs evaluate an entire generation and assign a single reward score at the end, whereas PRMs provide reward scores at each step of the generation, offering several notable advantages. PRMs enable fine-grained verification of LLM outputs, improving explainability by pinpointing errors at their exact location. This step-wise reward signaling also facilitates inference-time scaling, such as Monte Carlo Tree Search (MCTS), and supports step-by-step reinforcement learning \cite{wang2023math,snell2408scaling}. 

However, despite significant success in mathematical and coding tasks \cite{lightman2023let,wang2023math, ma2023let}, the application of PRMs to other domain remains underexplored \cite{beeching2024scalingtesttimecompute}. This gap arises from several key challenges: (1) \textbf{Verification of correctness}: Unlike mathematical or coding problems, which have objective ground-truth answers, verifying correctness in open-ended generative tasks is inherently more complex. (2) \textbf{Process-supervised data collection}: Training PRMs requires datasets with annotations for each step of the answer, which makes it difficult to scale when relying on human annotators. 

To address these challenges, we introduce a novel framework for training PRMs to verify open-ended text generation in the clinical domain (Figure~\ref{fig:overview}). We specifically target the rapidly growing use case of \textbf{ambient scribing}, where LLMs generate clinical notes from patient-doctor dialogues~\cite{barr2024preparing}. Drawing on domain expertise, we carefully design clinically meaningful ``steps'' and inject realistic ``errors'' to reflect authentic documentation practices. To scale supervision, we employ an LLM to generate process-labeled data by systematically introducing errors into high-quality reference notes. Our main contributions are as follows:

\begin{enumerate}[leftmargin=*]
    \item Our PRM, built on LLaMA-3.1 8B, significantly outperforms both proprietary reasoning and non-reasoning models for clinical note verification. It achieves 98.8\% accuracy in detecting erroneous samples and 56.2\% accuracy in selecting physician-preferred notes, setting a new SOTA in this critical domain.
    \item We broaden the scope of PRMs beyond domains with clear ground-truth answers, introducing a generalizable framework that integrates LLMs with domain knowledge to generate process-supervised data at scale.
    \item To support future research, we release our full dataset, including process supervision annotations and physician preference labels.
\end{enumerate}

\section{Background and Related Work}\label{background}
\paragraph{Verify LLM Outputs Step-by-Step} Recent research highlights a growing interest in process supervision of LLMs to improve their reasoning capabilities. In contrast to earlier work on ORMs \cite{cobbe2021training,yu2024ovm}, PRMs assess the correctness of LLM outputs step-by-step \cite{lightman2023let,uesato2022solving}. One of the core applications of PRMs is Best-of-N sampling, where N complete solutions are generated, and the most suitable one is selected based on the PRM's reward signal. This method has proven exceptionally effective in mathematical reasoning and code generation tasks, frequently surpassing ORMs \cite{wang2023math, ma2023let}. Furthermore, PRMs can facilitate step-level reinforcement learning and guide inference-time scaling through techniques such as MCTS \cite{wang2023math, ma2023let, snell2408scaling}. They are considered crucial to the success of advanced reasoning models \cite{jiang2024technical, qin2024o1, wang2024openr}.

\paragraph{Process-Supervised Data} Acquiring human-labeled annotations for each step of PRM training data poses a prohibitive scaling cost, a challenge central to our work. To address this limitation, several studies have proposed methods to automatically generate process supervision data, using techniques such as Monte Carlo estimation \cite{wang2023math, luo2024improve, wang2024multi}. However, these approaches are primarily confined to the mathematical domain, where they are effective due to the unambiguous and easily verifiable nature of mathematical solutions. Our exploration of PRM is also related to prior work on evaluating step-by-step reasoning chains \cite{hao2024llm}, though our focus is on open-ended text generation tasks, which present unique challenges.

\paragraph{Generate Clinical Note From Dialogue} Ambient scribe systems that transcribe patient-provider conversations into clinical notes are among the most impactful applications of LLMs in healthcare, offering substantial reductions in documentation burden~\cite{barr2024preparing, tierney2025ambient}. This task was explored at the 2023 ACL ClinicalNLP and CLEFImage workshops, where the best results were achieved using proprietary models \cite{abacha2023overview, yim2023overview}. Recent studies also show promise in fine-tuning open-source LLMs to generate expert-level clinical notes \cite{wang2024towards, wang2023notechat, li2024improving}, though challenges such as inaccuracies and hallucinations remain.

\section{Our Approach: Building PRMs to Verify Clinical Note Generation}
\label{method}
In this section, we first introduce the task, followed by an explanation of the core method for constructing the process supervision dataset. We then discuss the training, usage, and evaluation methods for PRMs. An overview of our study is illustrated in Figure \ref{fig:overview}.

\subsection{Problem Formation}
Given a patient-doctor dialogue, an LLM can generate clinical notes comparable to those written by physicians. We aim to train PRMs to effectively verify LLM-generated notes step-by-step, enabling them to evaluate multiple candidates and select the most accurate and physician-preferred note. Guided by recent work (see details in Appendix \ref{app:note_ap}), we target the most challenging ``Assessment and Plan'' (A\&P) section of the note.

\begin{figure}[!htb]
    \captionsetup{labelfont=bf={bf},skip=12pt}
    \centering
    \includegraphics[width=1\columnwidth]{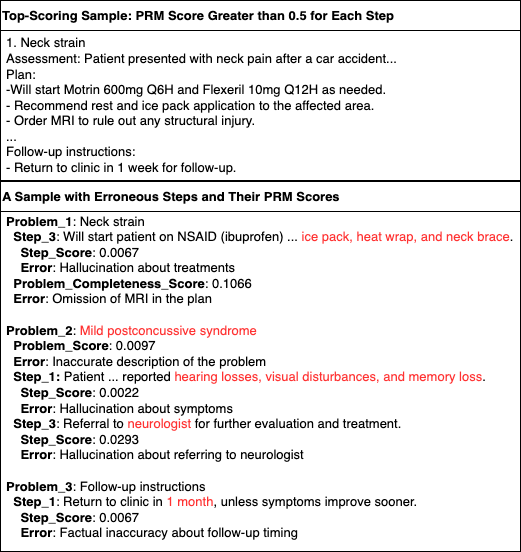}
    \caption{\textbf{Two notes to the same dialogue, graded by the PRM.} In the top-scoring sample, the note is accurate and concise, with the PRM score (probability of a ``+'' score for a given step) exceeding 0.5 for each step. In the negative sample, various errors occur across steps, and PRM effectively assigns low scores to those erroneous steps. Items ending in \textbf{\_Score} represent the PRM score assigned to each step. }
    \label{fig:case}
    \vspace{-10pt}
\end{figure}

\subsection{Training Data Construction}
\label{training}
\paragraph{Base Dataset} We utilized the Dialogue-G dataset and modified ACI-BENCH \cite{wang2024towards, yim2023aci}. These datasets comprise synthetic transcripts of patient-doctor dialogues and corresponding notes generated by Gemini Pro 1.0, strictly adhering to the ``Best Practice'' format recommended by a panel of internal medicine physicians. The original notes from these datasets are considered the \textit{gold-reference}.

\paragraph{Define Steps} Unlike mathematical problems, where each solution step is typically a single sentence, the concept of a ``step'' in an open-ended text generation task is less clear. The A\&P sections of clinical notes are semi-structured, consisting of multiple problems, each encompassing several clinical reasoning narratives related to their assessment and plan (Figure \ref{fig:overview} and \ref{fig:case}). While various design options exist, we adopted a heuristic approach informed by physician expertise (see Section~\ref{app:note_step} for details), as outlined below.


\noindent
\begin{enumerate}[leftmargin=*] 
\item The description of each problem is treated as a step, reflecting the importance of the problem list, which is critical for various medical coding and insurance purposes.
\item Within each problem, every sentence constitutes an individual step. 
\item After all sentences within each problem, we include a \textbf{problem-level completeness} step. Similarly, after addressing all problems, we add a \textbf{note-level completeness} step. This design aims to prevent the reward model from favoring shorter but incomplete answers. 
\item At the end of the note, we introduce an \textbf{end-of-note} step, with its score label representing the overall quality of the note. 
\end{enumerate}

\paragraph{Introduce Errors} Through discussions with clinical experts who reviewed LLM-generated notes, we identified three common types of errors:
\begin{itemize}[leftmargin=*]
\item \textbf{Factual Inaccuracy}: Errors involving incorrect information that are referenced in the conversation but not supported by its content.
\item \textbf{Hallucination}: Introduction of entirely unrelated subject entities that were not mentioned in the conversation.
\item \textbf{Unhelpfulness}: Expressions that are vague, incomplete, confusing, or lacking critical details.
\end{itemize}

We tasked Gemini Pro 1.5~\cite{team2024gemini} with generating a pool of unique errors for each error type in each case. This was achieved through careful prompt engineering and manual inspection of the errors by our physician co-authors. Subsequently, we randomly swapped original steps in the gold-reference notes with entries from the error pools, ensuring that each error was used only once. At this stage, we introduced a fourth error type, \textbf{Incompleteness}, by randomly removing specific steps or entire problems from the samples. 

\paragraph{Introduce Semantic Diversity} Lastly, we tasked Gemini Pro 1.5 with generating a pool of paraphrased sentences to enhance the semantic diversity of the notes. These paraphrases were designed to convey the same information as the original sentences without introducing any new content. We randomly replaced the remaining correct steps in the samples with these paraphrases.

\begin{table}[ht]
\centering
\captionsetup{labelfont=bf={bf},skip=12pt}
\renewcommand{\tablename}{Table}
\small
\begin{tabular}{ll}
\toprule
 & \textbf{Counts} \\
\midrule
Total No. of cases by data source &  \\
\quad ACI-BENCH & 67 \\
\quad Dialogue-G & 1205 \\
Total No. of samples & 9680 \\
Mean No. of samples per case & 7.61 \\
Mean No. of errors per sample &  \\
\quad Factual Inaccuracy & 1.16 \\
\quad Hallucination & 1.18 \\
\quad Unhelpfulness & 1.19 \\
\quad Incompleteness & 1.27 \\
Mean No. of paraphrases per sample & 2.38 \\
\bottomrule
\end{tabular}
\caption{\textbf{Summary statistics of PRM-Clinic.} The reported numbers are calculated solely from negative samples. Samples from the same case share the same dialogues. During training, gold-reference samples are randomly mixed with negative samples.}
\label{tab:summary_statistics}
\vspace{-10pt}
\end{table}

\paragraph{Step Scores}
\uline{Each step is assigned a \textbf{score label} of ``\(+\)'' or ``\(-\)''.} The ``\(+\)'' label represents a correct step, and the ``\(-\)'' label indicates an incorrect or incomplete step. For gold-reference samples, every step receives a ``\(+\)'' label. For samples containing errors, the erroneous step is assigned a score label of ``\(-\)''. For samples with incompleteness, we also assign the corresponding problem-level or note-level completeness score label to ``\(-\)''. Lastly, for any samples containing errors, we assign a score label of ``\(-\)'' to the end-of-note step.

\paragraph{PRM-Clinic Dataset} We named our final dataset PRM-Clinic and present its statistics in Table \ref{tab:summary_statistics}. Cases derived from Dialogue-G account for 95.1\% of the total instances. For ACI-BENCH, we utilized its training subset \cite{yim2023aci}, but with clinical notes generated using Gemini Pro 1.0 \cite{wang2024towards}.

\begin{table*}[ht]
    \centering
    \captionsetup{labelfont=bf={bf},skip=12pt}
    \renewcommand{\tablename}{Table}
     \renewcommand{\arraystretch}{1.5} 
    \small
    \begin{tabular}{p{1.5cm} p{2cm} p{2.2cm} p{1cm} p{1.25cm} p{5cm}}
        \toprule
        \textbf{Tasks} & \textbf{Reference Type} & \textbf{Data Source} & \textbf{Case Counts} & \textbf{Data Distribution} & \textbf{Notes} \\
        \midrule
        A-Prefer & Physician-preference & ACI-BENCH test2 and test3 subsets & 80 & OOD & Includes physician preference data used in the first round of RLHF in \cite{wang2024towards}. \uline{Each case contains three notes generated by an early phase of LLaMA-Clinic.} \\
        A-Verify & Gold-reference & ACI-BENCH test2 and test3 subsets & 80 & OOD & Gold-reference notes generated by an later phase of LLaMA-Clinic. Negative samples are introduced in the same manner as PRM-Clinic. \\
        Dialogue-G & Gold-reference & Dialogue-G testing subset & 80 & ID & Data constructed in the same manner as PRM-Clinic. \\
        A-Validate & Gold-reference & ACI-BENCH validation subset & 20 & ID & Data constructed in the same manner as PRM-Clinic. \\
        \bottomrule
    \end{tabular}
    \caption{\textbf{Details of the evaluation tasks.} For A-Prefer, A-Verify, and A-Validate, we utilized the original dialogues from the ACI-BENCH dataset. For ID tasks, the notes are generated by Gemini Pro, while for OOD tasks, the notes are generated by LLaMA-Clinic. ID: In-distribution. OOD: Out-of-distribution.}
    \label{tab:evaluation-dataset}
\end{table*}

\subsection{Training of PRMs} 
\paragraph{Training Objective}
The model is trained using a standard pretraining process to optimize the cross-entropy loss:
\[
L = -\sum_{i \in I} \log p_{\theta}(t_i \mid t_{<i}),
\]
where \(\{t_1, t_2, \dots, t_N\}\) denotes the sequence of tokens in the dataset. The model predicts the probability of the next token given the previous tokens, represented as \( p_{\theta}(t_i \mid t_{<i}) \), with \(\theta\) being the model parameters and \( t_{<i} = \{t_1, t_2, \dots, t_{i-1}\} \). The summation is taken over a set of token positions \( I \).

\paragraph{Training Details}
The training corpus consists of dialogue concatenated with clinical notes, where each step in the notes is separated by a step token followed by a score label token. We used LLaMA-3.1 8B Instruct \cite{dubey2024llama} as the base model, utilizing its reserved special tokens for all step and score label tokens. Further implementation details can be found in Section~\ref{app:implementation}.

\subsection{Usage of PRMs}
\label{section:usage}
\paragraph{Use PRMs at Inference Time} At inference time, we perform a single PRM forward pass over the entire clinical note to obtain PRM’s predicted \uline{score label} for each step. The clinical note itself is generated by a separate LLM. We define the \textbf{PRM score} as the probability that a given step is correct, which is the softmax probability of the special ``\(+\)'' token at that step position.

\paragraph{Use Model as PRM or ORM} To compare multiple notes, it is necessary to compute a single note-level score. \uline{A key implementation decision is whether to use the reward model as an \textbf{ORM}, which applies when the PRM score at the final end-of-note step serves as the note-level score.} In contrast, for PRM, the note-level score is computed as the product of all step-level PRM scores. We experimented with various scoring strategies and found that the product of step-level scores yielded the best results (Appendix \ref{app:scoring}).

\begin{figure*}[!ht]
    \captionsetup{labelfont=bf={bf},skip=12pt}
    \centering
    \includegraphics[width=1\linewidth]{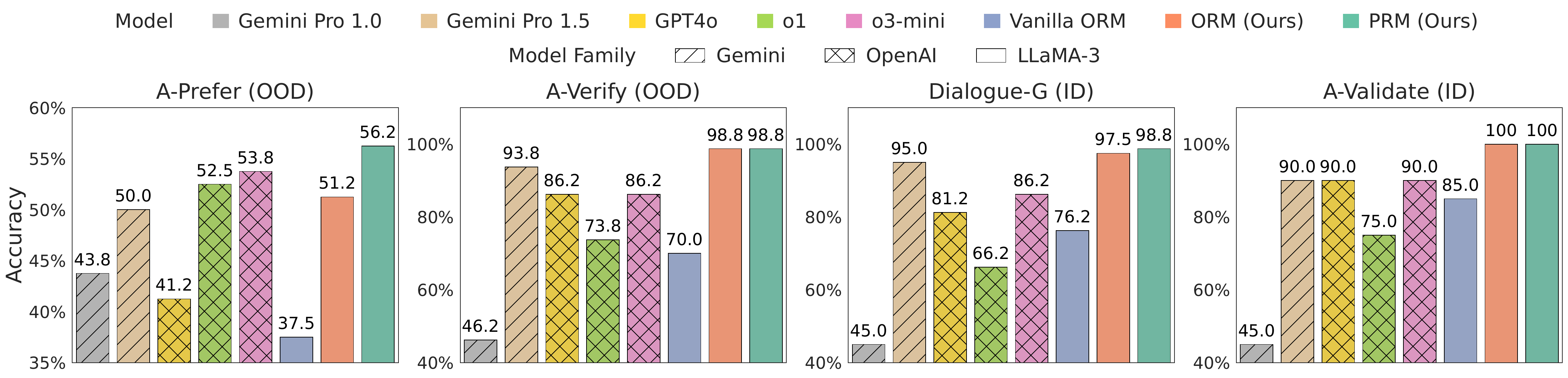}
    \caption{\textbf{Main results.} Our PRM outperforms all baseline models, including the vanilla ORM, Gemini Pro 1.5, and SOTA reasoning models, across all evaluated tasks. For A-Prefer, the task involves selecting the physician-preferred note from three candidate notes. For all other tasks, the objective is to identify the correct gold-reference sample from negative samples. For ID tasks, the notes are generated by Gemini Pro, while for OOD tasks, the notes are generated by LLaMA-Clinic. The results of the best-performing PRM are presented alongside the ORM results from the same checkpoint. ID: In-distribution. OOD: Out-of-distribution.}
    \label{fig:prm_results}
    \vspace{-5pt}
\end{figure*}

\subsection{Evaluation of PRMs}
\label{experiments}

\paragraph{Evaluation Metrics} We evaluated the reward model's ability to select the superior note within the Best-of-N framework using two key metrics: (1) \textbf{Accuracy of selecting the gold-reference note}, which measures the model's ability to identify the gold-reference note among error-containing samples (i.e., negative samples), and (2) \textbf{Accuracy of selecting the physician-preferred note}, which assesses its capability to identify the physician-preferred note from a set of candidate notes.

Notably, the task can be categorized as either in-distribution (ID) or out-of-distribution (OOD). Since our training dataset was developed using the Gemini Pro model family, tasks involving clinical notes generated by other models are considered OOD. For OOD tasks, we used outputs from LLaMA-Clinic. LLaMA-Clinic is a LLaMA-2 13B model that underwent comprehensive domain adaptation for clinical dialogue-to-note generation \cite{wang2024towards}. The details of the evaluation dataset are shown in Table~\ref{tab:evaluation-dataset}.

\paragraph{Baseline} We compare our PRM against several strong baselines, including LLM-as-a-judge setups using Gemini Pro 1.0 \cite{team2023gemini}, Gemini Pro 1.5 \cite{team2024gemini}, OpenAI's GPT-4o \cite{hurst2024gpt}, o1 \cite{jaech2024openai}, and o3-mini \cite{openai2025o3o4}. Additionally, we evaluate a vanilla ORM implementation, where the training corpus lacks step-level annotations and includes only a single score for the entire clinical note. For our ablation studies, we directly compare the performance of PRMs and ORMs derived from the same model checkpoint.

\paragraph{Physician Reader Study} To identify predictors of downstream Best-of-N performance, we conducted a comprehensive reader study involving nine physicians. For each case, LLaMA-Clinic~\cite{wang2024towards} generated 2,000 notes from the same patient-provider dialogue. We then used various PRM checkpoints to select the top-scoring note for physician evaluation. Each experiment was conducted in a randomized and blinded manner, involving at least three physicians. Additional details are provided in Section~\ref{app:reader}.

\begin{table*}[t!]
\centering
\linespread{1}
    \captionsetup{labelfont=bf={bf},skip=12pt}
    \renewcommand{\tablename}{Table}
\begin{adjustbox}{max width=\textwidth}
\begin{tabular}{@{} l cc cc cc cc @{}}
\toprule
\textbf{Model}  & \multicolumn{4}{c}{\textbf{Out-of-Distribution}}  & \multicolumn{4}{c}{\textbf{In-Distribution}}  \\
\cline{1-4}\cline{5-9}
 & \multicolumn{2}{c}{\textbf{A-Prefer}}  & \multicolumn{2}{c}{\textbf{A-Verify}}  & \multicolumn{2}{c}{\textbf{Dialogue-G}}  & \multicolumn{2}{c}{\textbf{A-Validate}}  \\
 & PRM & ORM  & PRM & ORM  & PRM & ORM  & PRM & ORM  \\
\midrule
\midrule
\multicolumn{9}{@{}l}{\textbf{Baseline (Vanilla Approach)}}\\
\quad All Data, Full-Token Loss & 46.2 & 38.8 & 95.0 & 97.5 & 98.8 & 100.0 & 100.0 & 100.0 \\
\midrule
\multicolumn{9}{@{}l}{\textbf{Ablation: Loss Functions}}\\
\quad Score-Token-Only Loss & 37.5 & 27.5 & 77.5 & 11.2 & 81.2 & 13.8 & 80.0 & 10.0 \\
\quad Special-Token Loss & 48.8 & 42.5 & 90.0 & 92.5 & 96.2 & 95.0 & 100.0 & 100.0 \\
\quad Notes-Only Loss & \textbf{55.0} & \textbf{48.8} & \textbf{91.2} & \textbf{96.2} & 96.2 & 98.8 & 100.0 & 100.0 \\
\midrule
\multicolumn{9}{@{}l}{\textbf{Ablation: Data Selection}}\\
\quad High Quality Only & \textbf{50.0} & \textbf{46.2} & 97.5 & 97.5 & 98.8 & 98.8 & 100.0 & 100.0 \\
\quad High Quality Only + Paraphrases & 47.5 & 36.2 & 95.0 & 93.8 & 97.5 & 96.2 & 100.0 & 95.0 \\
\quad High + Medium Quality & 45.0 & 42.5 & 86.2 & 97.5 & 95.0 & 98.8 & 100.0 & 100.0 \\
\quad High + Medium Quality + Paraphrases & 43.8 & 40.0 & \textbf{98.8} & \textbf{100.0} & 97.5 & 98.8 & 100.0 & 100.0 \\
\quad All Data + paraphrases & 43.8 & 40.0 & 97.5 & 98.8 & 97.5 & 98.8 & 100.0 & 100.0 \\
\midrule
\multicolumn{9}{@{}l}{\textbf{Notes-Only Loss with Data Selection}}\\
\quad High Quality Only & 35.0 & 32.5 & 87.5 & 96.2 & 95.0 & 98.8 & 100.0 & 100.0 \\
\quad High Quality Only + Paraphrases & 45.0 & 38.8 & 93.8 & 97.5 & 97.5 & 98.8 & 100.0 & 100.0 \\
\quad High + Medium Quality & 48.8 & 41.2 & 93.8 & 98.8 & 96.2 & 96.2 & 100.0 & 95.0 \\
\quad High + Medium Quality + Paraphrases & 46.2 & 42.5 & 96.2 & \textbf{100.0} & 97.5 & 98.8 & 95.0 & 100.0 \\
\quad All Data + Paraphrases & \textbf{56.2} & \textbf{51.2} & \textbf{98.8} & 98.8 & 98.8 & 97.5 & 100.0 & 100.0 \\
\bottomrule
\end{tabular}
\end{adjustbox}
\caption{\textbf{Ablation studies on loss functions and data selection.} We evaluated different strategies for cross-entropy loss computation and training data selection. The best performing model used notes-only loss and the full dataset with paraphrases. PRM and ORM results are reported from the checkpoint with the highest PRM A-Prefer performance. Numbers represent percentages of accuracy. \textbf{Bold} values indicate the highest PRM and ORM accuracy in each ablation study for OOD tasks. In ID tasks, multiple models achieved perfect accuracy.}
\label{tab:ablation}
\end{table*}

\section{Experimental Results}\label{results}

\subsection{Main Results}
\paragraph{Comparison with Baselines}
We present the main results in Figure \ref{fig:prm_results}. \textbf{Our PRM consistently outperforms all baselines across all tasks}, including the vanilla ORM, Gemini Pro 1.5, and SOTA reasoning models. Our analysis focuses on PRM performance in OOD settings (A-Verify and A-Prefer), given its near-perfect accuracy in the two ID verification tasks (Dialogue-G and A-Validate). Despite the inherent challenges of OOD tasks, PRM exhibits strong generalization, achieving 98.8\% accuracy on the A-Verify task (compared to 93.8\% for Gemini Pro 1.5). Among all tasks, A-Prefer proves to be the most challenging, reflecting the complexity of physician preferences. Here, PRM achieves 56.2\% accuracy in selecting physician-preferred notes (compared to 50.0\% for Gemini Pro 1.5). We discuss the explanation for the strong OOD performance in Section~\ref{app:rouge}. Importantly, while we utilized Gemini Pro 1.5 to generate synthetic data for PRM training, our results demonstrate that \textbf{PRMs significantly outperform Gemini Pro 1.5 itself}. This suggests that the proprietary model used for data generation doesn't impose an inherent upper bound on PRM performance within our methodology.

\paragraph{PRM vs. ORM}
Our model can function as either an ORM or a PRM, differing in the aggregation strategy from step-wise to note-level scoring, as detailed in Section \ref{section:usage}. In verification tasks, PRMs perform comparably to our ORMs when using the same model checkpoint, aligning with prior findings \cite{beeching2024scalingtesttimecompute, snell2408scaling}. However, in preference-based evaluations (A-Prefer), PRM surpasses our ORM, achieving 56.2\% accuracy compared to 51.2\%.

\paragraph{Comparison with Vanilla ORM}
Interestingly, our ORM implementation significantly outperforms the vanilla ORM, which predicts only the overall correctness of a generation without learning to estimate step-level reward signals. This suggests that even when functioning as an ORM and relying solely on the end-of-note score, \textbf{learning to predict step-level correctness during training is essential for achieving strong performance.}

\paragraph{Performance of Proprietary Models}
As expected, the most advanced reasoning models are better suited for the challenging A-Prefer task, with o3-mini delivering the strongest baseline performance. Notably, newer-generation models within each family show clear performance improvements (e.g., Gemini Pro 1.5 outperforms Gemini Pro 1.0, and o3-mini surpasses o1).

\subsection{Ablation Studies}
In the ablation studies, we examined loss functions and data selection strategies to identify key factors for PRM training. The results are presented in Table \ref{tab:ablation}.

\paragraph{Vanilla Implementation} In the vanilla implementation, the full PRM-Clinic dataset was used without the addition of paraphrases. The training objective is the standard cross-entropy loss applied to all tokens in the corpora, encompassing those from dialogues and notes. While the vanilla PRM model exhibits strong performance on the ID dataset, its performance is weaker on the OOD dataset.

\paragraph{Experiments with Loss Functions}
We further investigated the impact of various loss functions, all derived from the same cross-entropy loss framework but differing in the selection of tokens included in the loss computation. The formal definitions of these loss functions are detailed in Appendix \ref{appendix:loss function}. Restricting the loss computation to only the score label tokens resulted in a marked degradation of performance. Including losses over all special tokens (i.e., step tokens and score label tokens) produced performance similar to the vanilla approach but with slightly worse outcomes in verification tasks. 

Notably, masking out losses from dialogue tokens while restricting the loss computation to the notes led to a positive impact on performance, particularly in the A-Prefer task. These findings suggest that it is important for the PRM to learn to generate notes (based on the dialogue) while simultaneously learning to predict reward labels in a step-by-step manner. This observation may be influenced by the fact that our PRM did not undergo task-specific supervised finetuning prior to reward model training. Masking dialogue losses likely helps prevent overfitting, as dialogues often include lengthy, repetitive content across samples from the same case.

\paragraph{Experiments with Data Selection} While keeping the vanilla loss function, we first evaluated the impact of adding paraphrases to enhance semantic diversity. The results indicate that incorporating paraphrases is beneficial, leading to a general performance improvement in the A-Verify task compared to the baseline, although a slight decrease in A-Prefer accuracy was observed.

During manual inspection of the PRM-Clinic dataset, we identified lower-quality samples in the Dialogue-G subset, characterized by overly brief or unrealistic conversations. To address this, we experimented with selecting only cases with high or medium-quality dialogues, reflecting real-world conversations as judged by Gemini Pro. After filtering out low-quality data, the dataset contains 10,094 samples when retaining high and medium quality, and 5,854 samples when retaining only high quality, compared to the original dataset of 10,952 samples. Despite reducing the total number of training cases by up to 46\%, this selection process resulted in modest performance gains.

\paragraph{Best Results} 
Lastly, we experimented with different data selections while applying cross-entropy loss solely to the notes. Interestingly, the best results came from using the full dataset, which included all quality levels and added paraphrases. For the high-quality data group, masking out the dialogue loss caused a significant drop in performance. This suggests that when the training sample size is small, PRM may depend on learning from dialogue data to enhance step-level reward signal prediction. We used the best-performing PRM for the results shown in Figure \ref{fig:prm_results} and the physician reader study.

\begin{figure}[ht]
    \captionsetup{labelfont=bf={bf},skip=12pt}
    \centering
    \includegraphics[width=1\columnwidth]{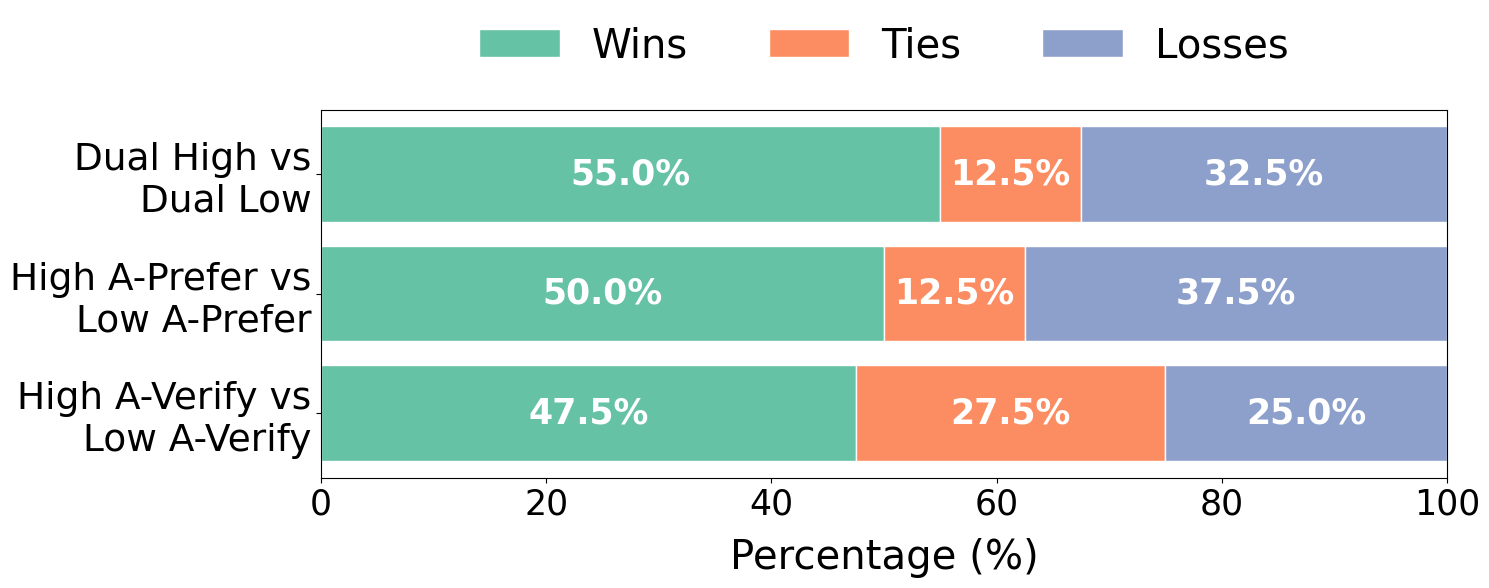}
    \caption{\textbf{Physician Reader Study Results.} We selected PRM checkpoints with varying A-Prefer and A-Verify performances for group comparisons. When analyzing the effect of a single metric, we identified two checkpoints with similar scores on the other metric. The Best-of-2000 note from each model was selected for physician review. Each group comparison involved at least three physicians, and the win rate was calculated based on the majority vote.}
    \label{fig:reader_study}
    \vspace{-10pt}
\end{figure}

\subsection{Physician Reader Study}
\label{app:reader}
In various ablation experiments, we frequently observed a divergence between the trends of A-Prefer (which selects physicians' preferred samples) and A-Verify (which selects gold-reference samples from those containing errors). To determine which metric better reflects the model's ability to select the highest-quality notes using Best-of-N in downstream tasks, we conducted a physician reader study. PRM checkpoints were selected based on varying combinations of A-Verify and A-Prefer performance (details provided in Appendix Table \ref{tab:reader study group}). The top-ranking notes from their Best-of-2000 selections were then submitted for physician review. As shown in Figure \ref{fig:reader_study}, the model with high performance on both metrics notably outperformed the one with low performance on both. Additionally, PRMs with high performance on either A-Prefer or A-Verify demonstrated superior note selection compared to those without, provided the other metric was of comparable value. This finding suggests that both metrics are predictive of Best-of-N performance. Notably, A-Prefer appears to have a greater impact than A-Verify, which is intuitively logical given that A-Prefer directly measures alignment with physician preferences.

\section{Conclusions and Future Work}
In this work, we introduce a methodology for developing PRMs for open-ended text generation in the clinical domain. Our PRM achieved SOTA performance in verifying clinical notes step-by-step and selecting physician-preferred notes via Best-of-N. Our work is among the first to demonstrate the effectiveness of PRMs for open-ended generative tasks beyond mathematics and coding. Notably, our framework is generalizable; it integrates LLMs and domain knowledge to produce process-supervised data at scale, and can be easily applied to other generative tasks across various domains.

While our PRM achieves the highest performance in predicting physician preferences, it is important to note that no current method achieves strong results. This suggests that additional factors influencing alignment with physicians remain unexplored, representing valuable opportunities for future research. Lastly, the application of PRMs for inference-time compute and scaling (e.g., MCTS) or step-by-step reinforcement learning is an exciting direction for future exploration.

\section{Acknowledgments}
This research was supported by NSF awards SCH-2205289. The funder played no role in the study design, data collection, analysis, and interpretation of data, or the writing of this manuscript. 

\section{Ethical Considerations}
Our work leveraged an LLM to generate synthetic data for training PRMs to verify clinical notes. In considering real-world deployment, several considerations are critical for minimizing harm in this clinical context. The scope of errors in synthetic data should be expanded beyond the four categories examined here to enable more comprehensive evaluation of clinical notes across diverse contexts. In addition, large-scale validation of PRM outputs is essential to identify unintended consequences such as overlooked errors or reward hacking. Finally, safeguards must be implemented to prevent synthetic data from introducing or amplifying biases, for instance when certain error types are disproportionately associated with specific disease populations, which could inadvertently reinforce inequities in documentation or care delivery.

\section{Limitations}
Our study is constrained by the limited number of physician reviewers and the relatively small dataset involved in the evaluation process. In addition, our approach to the task of clinical note generation builds upon prior work \cite{wang2024towards}, particularly their ``Best Practice'' note format, which was developed based on recommendations from a panel of internal medicine providers. When applying our framework to a different specialty, it may be necessary to recalibrate both the ``Best Practice'' note format and the corresponding definition of steps for the PRM. Lastly, our work used the proprietary Gemini model to generate synthetic data for PRM training, though the methodology could also be applied to other open-source frameworks (see Appendix \ref{app:open-source} for further discussion).

\bibliography{main}

\begin{thebibliography}{44}
\expandafter\ifx\csname natexlab\endcsname\relax\def\natexlab#1{#1}\fi

\bibitem[{Abacha et~al.(2023)Abacha, Yim, Adams, Snider, and Yetisgen-Yildiz}]{abacha2023overview}
Asma~Ben Abacha, Wen-wai Yim, Griffin Adams, Neal Snider, and Meliha Yetisgen-Yildiz. 2023.
\newblock Overview of the mediqa-chat 2023 shared tasks on the summarization \& generation of doctor-patient conversations.
\newblock In \emph{Proceedings of the 5th Clinical Natural Language Processing Workshop}, pages 503--513.

\bibitem[{Barr et~al.(2024)Barr, Gramling, and Vosoughi}]{barr2024preparing}
Paul~J Barr, Robert Gramling, and Soroush Vosoughi. 2024.
\newblock Preparing for the widespread adoption of clinic visit recording.
\newblock \emph{NEJM AI}, 1(11):AIp2400392.

\bibitem[{Beavins(2024)}]{beavins2024ai}
Emma Beavins. 2024.
\newblock \href {https://www.fiercehealthcare.com/ai-and-machine-learning/2025-outlook-whats-next-ai-scribes-and-virtual-care} {2025 outlook: What's next for ai scribes and virtual care}.
\newblock Fierce Healthcare, published December 18, 2024.

\bibitem[{Beeching et~al.()Beeching, Tunstall, and Rush}]{beeching2024scalingtesttimecompute}
Edward Beeching, Lewis Tunstall, and Sasha Rush.
\newblock \href {https://huggingface.co/spaces/HuggingFaceH4/blogpost-scaling-test-time-compute} {Scaling test-time compute with open models}.

\bibitem[{Cheng et~al.(2025)Cheng, Li, Xiong, Shao, and Lv}]{cheng2025pure}
Jie Cheng, Lijun Li, Gang Xiong, Jing Shao, and Yisheng Lv. 2025.
\newblock Pure: Prm is still effective and compute-efficient for llm math reasoning.
\newblock \url{https://github.com/CJReinforce/PURE}.

\bibitem[{Chowdhry et~al.(2017)Chowdhry, Mishuris, and Mann}]{chowdhry2017problem}
Shilpa~M Chowdhry, Rebecca~G Mishuris, and Devin Mann. 2017.
\newblock Problem-oriented charting: a review.
\newblock \emph{International journal of medical informatics}, 103:95--102.

\bibitem[{Cobbe et~al.(2021)Cobbe, Kosaraju, Bavarian, Chen, Jun, Kaiser, Plappert, Tworek, Hilton, Nakano et~al.}]{cobbe2021training}
Karl Cobbe, Vineet Kosaraju, Mohammad Bavarian, Mark Chen, Heewoo Jun, Lukasz Kaiser, Matthias Plappert, Jerry Tworek, Jacob Hilton, Reiichiro Nakano, et~al. 2021.
\newblock Training verifiers to solve math word problems.
\newblock \emph{arXiv preprint arXiv:2110.14168}.

\bibitem[{DeParle(2000)}]{deparle2000evaluation}
Nancy-Ann DeParle. 2000.
\newblock Evaluation \& management services guidelines.
\newblock \emph{JAMA}, 283(23):3061--3061.

\bibitem[{Dubey et~al.(2024)Dubey, Jauhri, Pandey, Kadian, Al-Dahle, Letman, Mathur, Schelten, Yang, Fan et~al.}]{dubey2024llama}
Abhimanyu Dubey, Abhinav Jauhri, Abhinav Pandey, Abhishek Kadian, Ahmad Al-Dahle, Aiesha Letman, Akhil Mathur, Alan Schelten, Amy Yang, Angela Fan, et~al. 2024.
\newblock The llama 3 herd of models.
\newblock \emph{arXiv preprint arXiv:2407.21783}.

\bibitem[{Guo et~al.(2025)Guo, Yang, Zhang, Song, Zhang, Xu, Zhu, Ma, Wang, Bi et~al.}]{guo2025deepseek}
Daya Guo, Dejian Yang, Haowei Zhang, Junxiao Song, Ruoyu Zhang, Runxin Xu, Qihao Zhu, Shirong Ma, Peiyi Wang, Xiao Bi, et~al. 2025.
\newblock Deepseek-r1: Incentivizing reasoning capability in llms via reinforcement learning.
\newblock \emph{arXiv preprint arXiv:2501.12948}.

\bibitem[{Hao et~al.(2024)Hao, Gu, Luo, Liu, Shao, Wang, Xie, Ma, Samavedhi, Gao et~al.}]{hao2024llm}
Shibo Hao, Yi~Gu, Haotian Luo, Tianyang Liu, Xiyan Shao, Xinyuan Wang, Shuhua Xie, Haodi Ma, Adithya Samavedhi, Qiyue Gao, et~al. 2024.
\newblock Llm reasoners: New evaluation, library, and analysis of step-by-step reasoning with large language models.
\newblock \emph{arXiv preprint arXiv:2404.05221}.

\bibitem[{Hurst et~al.(2024)Hurst, Lerer, Goucher, Perelman, Ramesh, Clark, Ostrow, Welihinda, Hayes, Radford et~al.}]{hurst2024gpt}
Aaron Hurst, Adam Lerer, Adam~P Goucher, Adam Perelman, Aditya Ramesh, Aidan Clark, AJ~Ostrow, Akila Welihinda, Alan Hayes, Alec Radford, et~al. 2024.
\newblock Gpt-4o system card.
\newblock \emph{arXiv preprint arXiv:2410.21276}.

\bibitem[{Jaech et~al.(2024)Jaech, Kalai, Lerer, Richardson, El-Kishky, Low, Helyar, Madry, Beutel, Carney et~al.}]{jaech2024openai}
Aaron Jaech, Adam Kalai, Adam Lerer, Adam Richardson, Ahmed El-Kishky, Aiden Low, Alec Helyar, Aleksander Madry, Alex Beutel, Alex Carney, et~al. 2024.
\newblock Openai o1 system card.
\newblock \emph{arXiv preprint arXiv:2412.16720}.

\bibitem[{Jiang et~al.(2024)Jiang, Chen, Min, Chen, Cheng, Wang, Tang, Sun, Deng, Zhao et~al.}]{jiang2024technical}
Jinhao Jiang, Zhipeng Chen, Yingqian Min, Jie Chen, Xiaoxue Cheng, Jiapeng Wang, Yiru Tang, Haoxiang Sun, Jia Deng, Wayne~Xin Zhao, et~al. 2024.
\newblock Technical report: Enhancing llm reasoning with reward-guided tree search.
\newblock \emph{arXiv preprint arXiv:2411.11694}.

\bibitem[{Jin et~al.(2024)Jin, Chen, Zhou, Xu, Cheung, Chen, Summers, Rousseau, Ni, Landsman et~al.}]{jin2024hidden}
Qiao Jin, Fangyuan Chen, Yiliang Zhou, Ziyang Xu, Justin~M Cheung, Robert Chen, Ronald~M Summers, Justin~F Rousseau, Peiyun Ni, Marc~J Landsman, et~al. 2024.
\newblock Hidden flaws behind expert-level accuracy of multimodal gpt-4 vision in medicine.
\newblock \emph{ArXiv}, pages arXiv--2401.

\bibitem[{Li et~al.(2018)Li, Garg, Cun, Shieh, Krishnan, Fang, and Chen}]{li2018impact}
Ron~C Li, Trit Garg, Tony Cun, Lisa Shieh, Gomathi Krishnan, Daniel Fang, and Jonathan~H Chen. 2018.
\newblock Impact of problem-based charting on the utilization and accuracy of the electronic problem list.
\newblock \emph{Journal of the American Medical Informatics Association}, 25(5):548--554.

\bibitem[{Li et~al.(2024)Li, Wu, Smith, Lo, and Liu}]{li2024improving}
Yizhan Li, Sifan Wu, Christopher Smith, Thomas Lo, and Bang Liu. 2024.
\newblock Improving clinical note generation from complex doctor-patient conversation.
\newblock \emph{arXiv preprint arXiv:2408.14568}.

\bibitem[{Lightman et~al.(2023)Lightman, Kosaraju, Burda, Edwards, Baker, Lee, Leike, Schulman, Sutskever, and Cobbe}]{lightman2023let}
Hunter Lightman, Vineet Kosaraju, Yura Burda, Harri Edwards, Bowen Baker, Teddy Lee, Jan Leike, John Schulman, Ilya Sutskever, and Karl Cobbe. 2023.
\newblock Let's verify step by step.
\newblock \emph{arXiv preprint arXiv:2305.20050}.

\bibitem[{Liu et~al.(2025)Liu, Gao, Zhao, Zhang, Li, Qi, Ouyang, and Zhou}]{liu2025can}
Runze Liu, Junqi Gao, Jian Zhao, Kaiyan Zhang, Xiu Li, Biqing Qi, Wanli Ouyang, and Bowen Zhou. 2025.
\newblock Can 1b llm surpass 405b llm? rethinking compute-optimal test-time scaling.
\newblock \emph{arXiv preprint arXiv:2502.06703}.

\bibitem[{Liu et~al.(2024)Liu, Hetherington, Dharod, Carroll, Bundy, Nguyen, Bundy, Isreal, McWilliams, and Cleveland}]{liu2024does}
Tsai-Ling Liu, Timothy~C Hetherington, Ajay Dharod, Tracey Carroll, Richa Bundy, Hieu Nguyen, Henry~E Bundy, McKenzie Isreal, Andrew McWilliams, and Jeffrey~A Cleveland. 2024.
\newblock Does ai-powered clinical documentation enhance clinician efficiency? a longitudinal study.
\newblock \emph{NEJM AI}, page AIoa2400659.

\bibitem[{Loshchilov and Hutter(2017)}]{loshchilov2017decoupled}
Ilya Loshchilov and Frank Hutter. 2017.
\newblock Decoupled weight decay regularization.
\newblock \emph{arXiv preprint arXiv:1711.05101}.

\bibitem[{Luo et~al.(2024)Luo, Liu, Liu, Phatale, Lara, Li, Shu, Zhu, Meng, Sun et~al.}]{luo2024improve}
Liangchen Luo, Yinxiao Liu, Rosanne Liu, Samrat Phatale, Harsh Lara, Yunxuan Li, Lei Shu, Yun Zhu, Lei Meng, Jiao Sun, et~al. 2024.
\newblock Improve mathematical reasoning in language models by automated process supervision.
\newblock \emph{arXiv preprint arXiv:2406.06592}.

\bibitem[{Ma et~al.(2023)Ma, Zhou, Liu, Yuan, Liu, You, and Yang}]{ma2023let}
Qianli Ma, Haotian Zhou, Tingkai Liu, Jianbo Yuan, Pengfei Liu, Yang You, and Hongxia Yang. 2023.
\newblock Let's reward step by step: Step-level reward model as the navigators for reasoning.
\newblock \emph{arXiv preprint arXiv:2310.10080}.

\bibitem[{Omiye et~al.(2024)Omiye, Gui, Rezaei, Zou, and Daneshjou}]{omiye2024large}
Jesutofunmi~A Omiye, Haiwen Gui, Shawheen~J Rezaei, James Zou, and Roxana Daneshjou. 2024.
\newblock Large language models in medicine: the potentials and pitfalls: a narrative review.
\newblock \emph{Annals of Internal Medicine}, 177(2):210--220.

\bibitem[{OpenAI(2025)}]{openai2025o3o4}
OpenAI. 2025.
\newblock \href {https://cdn.openai.com/pdf/2221c875-02dc-4789-800b-e7758f3722c1/o3-and-o4-mini-system-card.pdf} {Openai o3 and o4-mini system card}.
\newblock Accessed: 2025-05-19.

\bibitem[{Ouyang et~al.(2022)Ouyang, Wu, Jiang, Almeida, Wainwright, Mishkin, Zhang, Agarwal, Slama, Ray et~al.}]{ouyang2022training}
Long Ouyang, Jeffrey Wu, Xu~Jiang, Diogo Almeida, Carroll Wainwright, Pamela Mishkin, Chong Zhang, Sandhini Agarwal, Katarina Slama, Alex Ray, et~al. 2022.
\newblock Training language models to follow instructions with human feedback.
\newblock \emph{Advances in neural information processing systems}, 35:27730--27744.

\bibitem[{Qin et~al.(2024)Qin, Li, Zou, Liu, Xia, Huang, Ye, Yuan, Liu, Li et~al.}]{qin2024o1}
Yiwei Qin, Xuefeng Li, Haoyang Zou, Yixiu Liu, Shijie Xia, Zhen Huang, Yixin Ye, Weizhe Yuan, Hector Liu, Yuanzhi Li, et~al. 2024.
\newblock O1 replication journey: A strategic progress report--part 1.
\newblock \emph{arXiv preprint arXiv:2410.18982}.

\bibitem[{Rasley et~al.(2020)Rasley, Rajbhandari, Ruwase, and He}]{rasley2020deepspeed}
Jeff Rasley, Samyam Rajbhandari, Olatunji Ruwase, and Yuxiong He. 2020.
\newblock Deepspeed: System optimizations enable training deep learning models with over 100 billion parameters.
\newblock In \emph{Proceedings of the 26th ACM SIGKDD international conference on knowledge discovery \& data mining}, pages 3505--3506.

\bibitem[{Snell et~al.(2024)Snell, Lee, Xu, and Kumar}]{snell2408scaling}
Charlie Snell, Jaehoon Lee, Kelvin Xu, and Aviral Kumar. 2024.
\newblock Scaling llm test-time compute optimally can be more effective than scaling model parameters.
\newblock \emph{arXiv preprint arXiv:2410.09671}.

\bibitem[{Stiennon et~al.(2020)Stiennon, Ouyang, Wu, Ziegler, Lowe, Voss, Radford, Amodei, and Christiano}]{stiennon2020learning}
Nisan Stiennon, Long Ouyang, Jeffrey Wu, Daniel Ziegler, Ryan Lowe, Chelsea Voss, Alec Radford, Dario Amodei, and Paul~F Christiano. 2020.
\newblock Learning to summarize with human feedback.
\newblock \emph{Advances in Neural Information Processing Systems}, 33:3008--3021.

\bibitem[{Team et~al.(2023)Team, Anil, Borgeaud, Alayrac, Yu, Soricut, Schalkwyk, Dai, Hauth, Millican et~al.}]{team2023gemini}
Gemini Team, Rohan Anil, Sebastian Borgeaud, Jean-Baptiste Alayrac, Jiahui Yu, Radu Soricut, Johan Schalkwyk, Andrew~M Dai, Anja Hauth, Katie Millican, et~al. 2023.
\newblock Gemini: a family of highly capable multimodal models.
\newblock \emph{arXiv preprint arXiv:2312.11805}.

\bibitem[{Team et~al.(2024)Team, Georgiev, Lei, Burnell, Bai, Gulati, Tanzer, Vincent, Pan, Wang et~al.}]{team2024gemini}
Gemini Team, Petko Georgiev, Ving~Ian Lei, Ryan Burnell, Libin Bai, Anmol Gulati, Garrett Tanzer, Damien Vincent, Zhufeng Pan, Shibo Wang, et~al. 2024.
\newblock Gemini 1.5: Unlocking multimodal understanding across millions of tokens of context.
\newblock \emph{arXiv preprint arXiv:2403.05530}.

\bibitem[{Tierney et~al.(2025)Tierney, Gayre, Hoberman, Mattern, Ballesca, Wilson~Hannay, Castilla, Lau, Kipnis, Liu et~al.}]{tierney2025ambient}
Aaron~A Tierney, Gregg Gayre, Brian Hoberman, Britt Mattern, Manuel Ballesca, Sarah~B Wilson~Hannay, Kate Castilla, Cindy~S Lau, Patricia Kipnis, Vincent Liu, et~al. 2025.
\newblock Ambient artificial intelligence scribes: learnings after 1 year and over 2.5 million uses.
\newblock \emph{NEJM Catalyst Innovations in Care Delivery}, 6(5):CAT--25.

\bibitem[{Uesato et~al.(2022)Uesato, Kushman, Kumar, Song, Siegel, Wang, Creswell, Irving, and Higgins}]{uesato2022solving}
Jonathan Uesato, Nate Kushman, Ramana Kumar, Francis Song, Noah Siegel, Lisa Wang, Antonia Creswell, Geoffrey Irving, and Irina Higgins. 2022.
\newblock Solving math word problems with process-and outcome-based feedback.
\newblock \emph{arXiv preprint arXiv:2211.14275}.

\bibitem[{Van~Veen et~al.(2024)Van~Veen, Van~Uden, Blankemeier, Delbrouck, Aali, Bluethgen, Pareek, Polacin, Reis, Seehofnerov{\'a} et~al.}]{van2024adapted}
Dave Van~Veen, Cara Van~Uden, Louis Blankemeier, Jean-Benoit Delbrouck, Asad Aali, Christian Bluethgen, Anuj Pareek, Malgorzata Polacin, Eduardo~Pontes Reis, Anna Seehofnerov{\'a}, et~al. 2024.
\newblock Adapted large language models can outperform medical experts in clinical text summarization.
\newblock \emph{Nature medicine}, 30(4):1134--1142.

\bibitem[{Wang et~al.(2024{\natexlab{a}})Wang, Gao, Liu, Xu, Hussein, Labban, Iheasirim, Korsapati, and Sun}]{wang2024towards}
Hanyin Wang, Chufan Gao, Bolun Liu, Qiping Xu, Guleid Hussein, Mohamad~El Labban, Kingsley Iheasirim, Hariprasad Korsapati, and Jimeng Sun. 2024{\natexlab{a}}.
\newblock Towards adapting open-source large language models for expert-level clinical note generation.
\newblock \emph{arXiv preprint arXiv:2405.00715}.

\bibitem[{Wang et~al.(2024{\natexlab{b}})Wang, Fang, Wan, Wen, Zhu, Liu, Gong, Song, Chen, Ni et~al.}]{wang2024openr}
Jun Wang, Meng Fang, Ziyu Wan, Muning Wen, Jiachen Zhu, Anjie Liu, Ziqin Gong, Yan Song, Lei Chen, Lionel~M Ni, et~al. 2024{\natexlab{b}}.
\newblock Openr: An open source framework for advanced reasoning with large language models.
\newblock \emph{arXiv preprint arXiv:2410.09671}.

\bibitem[{Wang et~al.(2023{\natexlab{a}})Wang, Yao, Yang, Zhou, Li, Wang, Xu, and Yu}]{wang2023notechat}
Junda Wang, Zonghai Yao, Zhichao Yang, Huixue Zhou, Rumeng Li, Xun Wang, Yucheng Xu, and Hong Yu. 2023{\natexlab{a}}.
\newblock Notechat: a dataset of synthetic doctor-patient conversations conditioned on clinical notes.
\newblock \emph{arXiv preprint arXiv:2310.15959}.

\bibitem[{Wang et~al.(2023{\natexlab{b}})Wang, Li, Shao, Xu, Dai, Li, Chen, Wu, and Sui}]{wang2023math}
Peiyi Wang, Lei Li, Zhihong Shao, RX~Xu, Damai Dai, Yifei Li, Deli Chen, Y~Wu, and Zhifang Sui. 2023{\natexlab{b}}.
\newblock Math-shepherd: A label-free step-by-step verifier for llms in mathematical reasoning.
\newblock \emph{arXiv preprint arXiv:2312.08935}.

\bibitem[{Wang et~al.(2024{\natexlab{c}})Wang, Li, Wu, Luo, Hou, Yu, and Shang}]{wang2024multi}
Zihan Wang, Yunxuan Li, Yuexin Wu, Liangchen Luo, Le~Hou, Hongkun Yu, and Jingbo Shang. 2024{\natexlab{c}}.
\newblock Multi-step problem solving through a verifier: An empirical analysis on model-induced process supervision.
\newblock \emph{arXiv preprint arXiv:2402.02658}.

\bibitem[{Wolf et~al.(2019)Wolf, Debut, Sanh, Chaumond, Delangue, Moi, Cistac, Rault, Louf, Funtowicz et~al.}]{wolf2019huggingface}
Thomas Wolf, Lysandre Debut, Victor Sanh, Julien Chaumond, Clement Delangue, Anthony Moi, Pierric Cistac, Tim Rault, R{\'e}mi Louf, Morgan Funtowicz, et~al. 2019.
\newblock Huggingface's transformers: State-of-the-art natural language processing.
\newblock \emph{arXiv preprint arXiv:1910.03771}.

\bibitem[{Yim et~al.(2023{\natexlab{a}})Yim, Abacha, Snider, Adams, and Yetisgen}]{yim2023overview}
Wen-wai Yim, A~Ben Abacha, N~Snider, G~Adams, and Meliha Yetisgen. 2023{\natexlab{a}}.
\newblock Overview of the mediqa-sum task at imageclef 2023: Summarization and classification of doctor-patient conversations.
\newblock In \emph{CLEF}.

\bibitem[{Yim et~al.(2023{\natexlab{b}})Yim, Fu, Ben~Abacha, Snider, Lin, and Yetisgen}]{yim2023aci}
Wen-wai Yim, Yujuan Fu, Asma Ben~Abacha, Neal Snider, Thomas Lin, and Meliha Yetisgen. 2023{\natexlab{b}}.
\newblock Aci-bench: a novel ambient clinical intelligence dataset for benchmarking automatic visit note generation.
\newblock \emph{Scientific Data}, 10(1):586.

\bibitem[{Yu et~al.(2024)Yu, Gao, and Wang}]{yu2024ovm}
Fei Yu, Anningzhe Gao, and Benyou Wang. 2024.
\newblock Ovm, outcome-supervised value models for planning in mathematical reasoning.
\newblock In \emph{Findings of the Association for Computational Linguistics: NAACL 2024}, pages 858--875.

\end{thebibliography}

\appendix
\setcounter{table}{0}
\setcounter{figure}{0}

\clearpage
\section{Formal Definitions of Loss Functions for Ablation Studies} \label{appendix:loss function}

\subsection{Vanilla Loss}
The vanilla implementation computes the cross-entropy loss over all tokens in the dataset, including dialogue and note tokens:
\[
L_{\text{vanilla}} = -\sum_{i \in I_{\text{all}}} \log p_{\theta}(t_i \mid t_{<i}),
\]
where $I_{\text{all}}$ represents the positions of all tokens in the sequence.

\subsection{Loss over Score Label Tokens Only}
Focusing on score label tokens only, the loss is computed as:
\[
L_{\text{score}} = -\sum_{i \in I_{\text{score}}} \log p_{\theta}(t_i \mid t_{<i}),
\]
where $I_{\text{score}}$ denotes the positions of score label tokens.

\subsection{Loss over All Special Tokens}
When including all special tokens, including both step tokens and score label tokens, the loss is defined as:
\[
L_{\text{special}} = -\sum_{i \in I_{\text{special}}} \log p_{\theta}(t_i \mid t_{<i}),
\]
with $I_{\text{special}}$ being the set of positions corresponding to special tokens.

\subsection{Loss over Notes Only}
Restricting the loss to note tokens by masking out dialogue tokens results in:
\[
L_{\text{notes}} = -\sum_{i \in I_{\text{notes}}} \log p_{\theta}(t_i \mid t_{<i}),
\]
where $I_{\text{notes}}$ represents the positions of note tokens, which also includes all special tokens.

\section{Details on Implementation}
\label{app:implementation}
\subsection{Training of PRM}
All models were trained using either four 80GB NVIDIA A100 or H100 GPUs, employing DeepSpeed ZeRO Stage 3 optimization~\cite{rasley2020deepspeed} and the AdamW optimizer~\cite{loshchilov2017decoupled}. We adopted Huggingface's Transformers library and utilized its Trainer module~\cite{wolf2019huggingface}. A limited search was conducted over learning rates [3e-4, 1.5e-4, 3e-5], with 3e-5 selected as the optimal value. The training process utilized consistent hyperparameters across all models, including a global batch size of 16, BF16 precision training, a warm-up phase of 50 steps, and a linear learning rate scheduler. All experiments were trained using one epoch of data.

\subsection{Training of Vanilla ORM}
\label{app:Vanilla}
For the training of vanilla ORM, we replace the true step score labels in the training corpora with the same placeholder token used during inference, except for the score label at the final end-of-note step. This effectively removes all step-level reward signals, retaining only a single score label for the entire generation during training. Otherwise, we keep the training corpora and pipeline the same as in the rest of the experiments.

\subsection{Use PRMs at Inference Time}
At inference time, we first use a regular expression to transform the LLM-generated clinical notes into the JSON format defined by our steps. We replace score label tokens with a placeholder token when concatenating the note with the dialogue to form the input context. We then perform a standard forward pass with the input context and obtain the softmax probability of the special ``\(+\)'' token at each step position, which represents the \textbf{PRM score} for that step.

When calculating the note-level score, we handle ORM and PRMs differently. For ORM, we simply use the PRM score at the final end-of-note step as the note-level score. For PRMs, to ensure mathematical stability, we avoid directly computing the product of step-level PRM scores. Instead, we take the logarithm of each score and sum the log values across all steps.

\subsection{Evaluation Metrics} \label{app:evalucation metrics}
We calculate accuracy for all metrics at the case level, where each case includes one gold-reference or physician-preferred sample and multiple negative samples. In verification tasks (A-Verify, A-Validation, and Dialogue-G), accuracy is determined by whether the top-scoring sample from the Best-of-N selection, based on PRM or ORM, matches the gold-reference sample. Similarly, in the A-Prefer task, accuracy is determined by whether the top-scoring sample from the Best-of-N selection, based on PRM or ORM, matches the physician-preferred sample.

\subsection{Physician Reader Study}
The physician reader study was performed using dialogues from ACI-BENCH test1 subset  (n = 40). We provided model details for each comparison group in Appendix Table \ref{tab:reader study group}. Checkpoints were selected from ablation studies to enable a controlled comparison of the effects of A-Prefer versus A-Verify.

When selecting top-scoring samples for physician review, we first attempt to select those in which the probability of each step's ``\(+\)'' score label exceeds that of its ``\(-\)'' score label. This approach mimics the strategy used in \cite{lightman2023let}. If no such samples exist, we select the top-scoring sample from all candidates.

\begin{table}[ht]
\centering
\renewcommand{\tablename}{Appendix Table}
\captionsetup{labelfont=bf, skip=12pt}
\resizebox{\columnwidth}{!}{%
\begin{tabular}{l c c}
\toprule
\textbf{Model (Checkpoint)} & \textbf{A-Prefer} & \textbf{A-Verify} \\ 
\midrule
\textbf{Dual High vs Dual Low} & & \\ 
\quad All Data + Paraphrases (500) & 56.2 & 98.8 \\ 
\quad Score-Token-Only Loss (685) & 33.8 & 86.2 \\ 
\textbf{High A-Prefer vs Low A-Prefer} & & \\ 
\quad Notes-Only Loss (685) & 55.0 & 91.2 \\ 
\quad Special-Token Loss (685) & 43.8 & 91.2 \\ 
\textbf{High A-Verify vs Low A-Verify} & & \\ 
\quad High Quality Only + Paraphrases (366) & 45.0 & 96.2 \\ 
\quad High + Medium Quality (631) & 45.0 & 86.2 \\ 
\bottomrule
\end{tabular}%
}
\caption{\textbf{Details of models used in the physician reader study.} Numbers represent percentages of accuracy from PRM.}
\label{tab:reader study group}
\end{table}

Of interest, the ``Best Practice'' note template was recommended by a panel of internal medicine physicians \cite{wang2024towards}. We explored collecting preferences from physicians in other specialties, such as obstetrics/gynecology and emergency medicine. As anticipated, there was a high rate of disagreement compared to the preferences of internal medicine physicians, particularly regarding fine-grained aspects such as the perceived target audience (e.g., peer physicians vs. patients) and the desired note length (e.g., concise vs. comprehensive). These findings underscore the complexity of physician preferences and highlight the importance of aligning priorities, criteria, and definitions of ``ideal'' documentation for real-world deployment. In this work, we report results exclusively from practicing internal medicine physicians.

\section{Additional Results}
\subsection{Distribution of Top-Scoring Samples}

Previous studies in the clinical domain have commonly employed lower temperature settings for generative tasks \cite{van2024adapted, wang2024towards}. In our study, we analyzed the distribution of top-scoring samples derived from PRMs. Specifically, when using LLaMA-Clinic to generate 2,000 samples per case for the physician reader study, we explored various temperature settings. For temperatures in the range of [0.2, 0.4, 0.6, 0.8, and 1.0], we generated 200 samples each. For temperatures ranging from 1.1 to 2.0, samples were generated at 0.1 intervals, with 100 samples per interval.

The distribution of the top-10 scoring samples across all 40 cases is illustrated in Appendix Figure \ref{fig:distribution}. Consistent with prior research, the most frequent temperature range for high-scoring samples was found to be between 0.4 and 1.0.

\begin{figure}[ht]
    \captionsetup{labelfont=bf={bf},skip=12pt}
    \centering
    \renewcommand{\figurename}{Appendix Figure}
    \includegraphics[width=1\columnwidth]{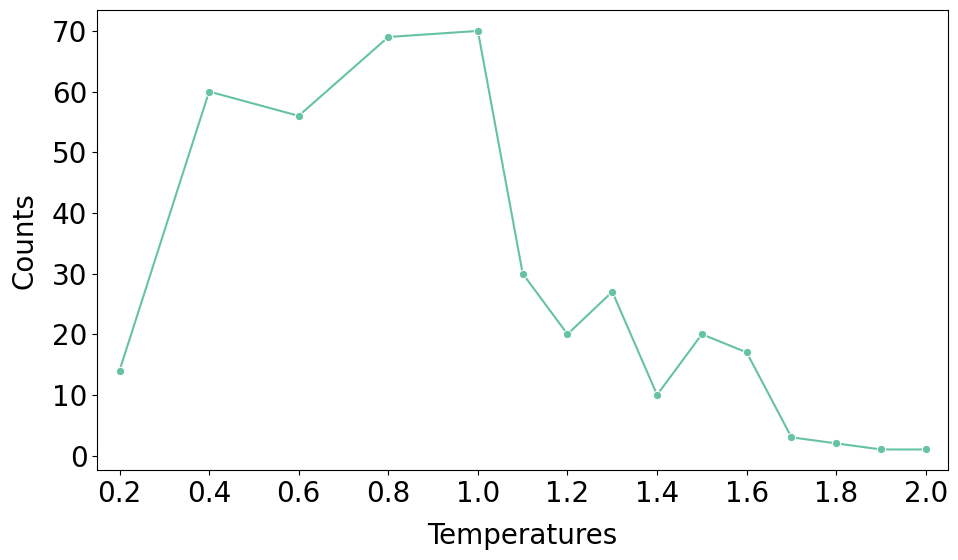}
    \caption{\textbf{Distribution of top-10 scoring samples by PRM across temperature settings.} This figure shows the counts of top-scoring samples per temperature setting for the physician reader study, with 2,000 samples generated using LLaMA-Clinic at various temperatures for each case. The highest frequency of top-scoring samples occurred in the temperature range of 0.4 to 1.0.}
    \label{fig:distribution}
\end{figure}

\subsection{Scoring Strategies} \label{app:scoring}
We present the A-Prefer results using various scoring strategies for note-level PRM scores, as shown in the Appendix Table \ref{tab:scoring_strategies}. These results are derived from the best-performing PRM. The highest performance was achieved by computing the note-level score as the product of all step-level PRM scores.

\begin{table*}[ht]
\centering
\renewcommand{\tablename}{Appendix Table}
\captionsetup{labelfont=bf}
\small
\begin{tabular}{ccccccccc}
\toprule
\textbf{Product} & \textbf{Last} & \textbf{Min} & \textbf{Mean} & \textbf{Median} & \textbf{Max} & \textbf{Geo Mean} & \textbf{Sum} & \textbf{Threshold} \\
\midrule
56.2\% & 51.2\% & 45.0\% & 55.0\% & 46.2\% & 42.5\% & 50.0\% & 37.5\% & 38.8\% \\
\bottomrule
\end{tabular}
\caption{\textbf{Accuracy of various scoring strategies for A-Prefer.} Each column represents a different method for deriving note-level scores from step-level PRM scores. The \textbf{Product} method multiplies all step-level PRM scores, utilizing the sum of their logarithms during calculations for mathematical stability. The \textbf{Last} method uses the step-level score at the end-of-note token, which is the approach used to obtain scores for ORM. \textbf{Min} takes the minimum score, \textbf{Mean} computes the average, \textbf{Median} uses the median, \textbf{Max} selects the maximum, \textbf{Geo Mean} calculates the geometric mean, \textbf{Sum} adds all scores, and \textbf{Threshold} counts the number of step-level PRM scores that exceed 0.5. Accuracy is determined by whether the top-scoring sample from each method matches the physician-preferred sample. Results are based on the best-performing PRM.}
\label{tab:scoring_strategies}
\end{table*}

\subsection{Quantitative Assessment of In-Distribution and Out-of-Distribution Clinical Notes}
\label{app:rouge}
To investigate the factors contributing to the robust performance of PRMs in out-of-distribution tasks (A-Prefer and A-Verify), we computed the ROUGE scores of clinical notes generated by physicians, LLaMA-Clinic, and Gemini Pro, based on prior work \cite{wang2024towards} (Appendix Table \ref{tab:rouge}). The high ROUGE scores across various cross-comparison groups suggest strong semantic similarities among all three sources. This result is unsurprising, given that different models should conform to the same best-practice standards and produce similarly high-quality notes for the same case.

\begin{table}[ht]
\centering
\renewcommand{\tablename}{Appendix Table}
\captionsetup{labelfont=bf, skip=12pt}
\resizebox{\columnwidth}{!}{%
\label{tab:note_comparison}
\begin{tabular}{lccc}
\toprule
\textbf{Model Comparison} & \textbf{ROUGE-1} & \textbf{ROUGE-L} & \textbf{ROUGE-Lsum} \\
\midrule
LLaMA-Clinic vs Physician & 0.46 & 0.34 & 0.43 \\
Gemini vs Physician & 0.49 & 0.37 & 0.46 \\
LLaMA-Clinic vs Gemini & 0.52 & 0.40 & 0.49 \\
\bottomrule
\end{tabular}
}
\caption{\textbf{Comparison of ROUGE scores among clinical Notes by physicians, LLaMA-Clinic, and Gemini Pro 1.0.} Cases are from the physician reader study in \cite{wang2024towards} and include only the ``Assessment and Plan'' section. Overall, high ROUGE scores were observed for each group comparison.}
\label{tab:rouge}
\end{table}

\section{Additional Discussion}
\subsection{Application of PRMs in Real-World Deployment}
\label{app:real-world}
Currently, no automated, scalable method exists to evaluate the quality of LLM-generated clinical notes at a fine-grained level. As a result, healthcare institutions rely on costly and time-consuming manual clinician evaluation as the gold standard. Taking the field of ambient AI scribes as an example, a healthcare institution may need to evaluate offerings from multiple industry vendors concurrently. Each evaluation group typically consists of tens to hundreds of clinicians who manually grade LLM-generated clinical notes and provide feedback based on a comprehensive rubric, primarily focusing on detecting errors, inaccuracies, or hallucinations. Furthermore, whenever vendors update their products—for instance, by incorporating a newer version of a foundation model—such evaluations must be repeated.

By applying effectively trained and validated PRMs, the evaluation of LLM-generated notes can be automated at scale, significantly reducing reliance on manual assessment. PRMs enable fine-grained, step-by-step verification of LLM outputs, precisely identifying errors at their exact locations. As demonstrated in our study, PRMs can also select notes that best align with physician preferences. Therefore, PRM-assessed performance can serve as a benchmark for evaluation, akin to the LLM-as-a-judge approach. Furthermore, our study shows that PRMs outperform the vanilla LLM-as-a-judge method. PRMs can also be employed for product monitoring after deployment in healthcare institutions, aiding in the detection of quality issues related to model shift and drift. Importantly, PRMs should first be validated on real-world data and undergo iteration and improvement—for example, through the introduction of additional error types.

Lastly, AI companies can use PRMs to enhance their products. This can be achieved through a simple Best-of-N filtering approach at inference time—similar to our experiments—ensuring that only the highest-quality note is output. Further model improvements are also promising with inference-time scaling techniques such as Monte Carlo Tree Search or step-wise reinforcement learning.

\subsection{Training PRMs with an Open-Source Framework}
\label{app:open-source}
The core of our methodology lies in designing ``errors'' and structuring ``steps'' to align with real-world clinical documentation needs. While we opted to use the proprietary Gemini model to generate synthetic errors due to its strong performance and ease of API access, our methodology is model-agnostic. These synthetic errors could just as effectively be generated using other powerful open-source models, such as LLaMA-3.1 405B \cite{dubey2024llama} or Deepseek-R1 \cite{guo2025deepseek} (given adequate GPU resources), provided that the critical decisions regarding error types are made beforehand. Furthermore, our model training is based on the open-source LLaMA model. Lastly, while we used Gemini Pro to generate synthetic cases for PRM development, PRMs outperform both Gemini Pro 1.0 and 1.5, indicating that the proprietary model does not impose an upper bound on PRM performance within our methodology.

\subsection{Assessment and Plan of Clinical Note}
\label{app:note_ap}
Building on recent research on ambient clinical charting and the real-world feedback \cite{wang2024towards}, we concentrate solely on the ``Assessment and Plan'' (A\&P) section of the clinical note in our work for the following reasons: (1) The A\&P section is arguably the most critical and time-intensive part of a clinical note, as it encapsulates key clinical reasoning and medical decision-making. (2) After domain-specific adaptation, the ``Subjective'' section produced by LLMs is generally of high quality and nearly indistinguishable from human-written content, whereas notable gaps persist in the A\&P section. (3) Other sections are considered less significant, as they are often imported directly from the electronic medical record (e.g., ``Objective'') or fit less naturally into the physician’s workflow during ambient scribing (e.g., ``Physical Exam'').

\subsection{Considerations in defining "steps" for clinical note}
\label{app:note_step}
When defining the "steps" for clinical note, we took into account the diverse real-world requirements of clinical documentation from various stakeholders. For example, the problem list within a note plays a critical role in billing processes for insurance companies and also serves as a key reference for daily clinical communication, warranting special attention~\cite{chowdhry2017problem}. In addition, we drew on existing guidelines and best practices for clinical note writing, which explicitly emphasize the value of structured, problem-oriented documentation~\cite{, li2018impact, deparle2000evaluation}.

\section{Additional Related Work}
\label{app:additiona_related_work}
After our preprint, DeepSeek-R1-Zero and DeepSeek-R1 were released \cite{guo2025deepseek}. These models demonstrate remarkable reasoning capabilities through large-scale reinforcement learning with rule-based rewards. In their paper, the authors discussed their reasons for not using PRMs, citing concerns such as reward hacking and the complexity of scaling.

In this exciting new paradigm of large-scale RL, we believe that PRMs remain effective for the following reasons: (1) As a simple yet powerful approach, PRMs serve as an effective means for Best-of-N filtering, guiding test-time computation. (2) In general domains where a verifiable answer is difficult to define, PRMs provide a reliable single reward signal when functioning as ORMs. As demonstrated in our study, this approach outperforms the standard ORM implementation. Moreover, recent research highlights the potential of PRMs in guiding test-time scaling, offering both effectiveness and computational efficiency \cite{liu2025can, cheng2025pure}.

\section{Prompts and Instructions}
We present the prompts to Geimino Pro for clinical notes-to-JSON format transformation (Appendix Table \ref{tab:prompt-json}), error generation (Appendix Table \ref{tab:promt-error}), paraphrase generation (Appendix Table \ref{tab:prompt-paraphrase}), case quality annotation (Appendix Table \ref{tab:prompt-eval}). and preferred note selection (Appendix Table \ref{tab:preferece-selection}) here. We provided instructions to reviewers in Appendix Table \ref{tab:reviewer-instruction}.

\section{PRM Examples}
We presented additional example notes along with their PRM scores in Appendix Table \ref{tab:prm-example1} and \ref{tab:prm-example2}.

\renewcommand{\arraystretch}{1.2}
\begin{table*}[ht]
    \small
    \captionsetup{labelfont=bf}
    \renewcommand{\tablename}{Appendix Table}

    \begin{tabular}{p{2.2cm} p{13cm}}
      
    \toprule
    \textbf{Category} & \textbf{Prompt} \\ 
    \midrule

    \textit{Note to Json} & Task: Given a clinical note, extract specific information and structure it into a JSON format according to the rules and schema provided.
    \newline
    \newline
    Instructions: \newline
    1. Output Format: \newline
    - The response should be in JSON format following the schema outlined below. \newline

    2. Sections to Ignore: \newline
    - Do not include any content labeled or titled ``Assessment and Plan''. \newline

    3. Content Extraction: \newline
    - Extract the remaining content of the clinical note and organize it into different problems. \newline
    - Problems can be identified using a numbered problem list within the note. \newline
    - When there's a section titled ``Follow-up instructions:'', treat this as a separate problem. \newline

    4. For Each Problem: \newline
    - Extract each step as a separate sentence. \newline
    - Ignore bullet point symbols, such as -, •, or other similar characters, when extracting steps. \newline
    - If the words ``Assessment:'' or ``Plan:'' appear in the clinical note, include those words at the beginning of the first sentence that follows their occurrence. \newline

    5. Adding Ratings: \newline
    - For each step, add a field called ``Step\_score'' with the value ``+''. \newline
    - For each problem, add a field called ``Problem\_score'' and ``Problem\_completeness\_score'' with the value ``+''. \newline

    6. Add Sequential Numbering: \newline
    - For each problem, add a field called ``Problem\_no''. Number them sequentially starting from ``1'', and use strings (e.g., ``"1"'', ``"2"''). \newline
    - For each step, add a field called ``Step\_no''. Number them sequentially starting from ``1'', and use strings. \newline

    7. Add Note Score: \newline
    - Add a field called ``Note\_completeness\_score'' with the value ``+'' at the root level of the JSON. \newline

    8. JSON Schema: \newline

    \{
    ``Problems'': [
      \{
        ``Problem'': ``Problem Description'',
        ``Problem\_no'': ``1'',
        ``Problem\_score'': ``+'',
        ``Steps'': [
          \{
            ``Step'': ``First step of the problem.'',
            ``Step\_no'': ``1'',
            ``Step\_score'': ``+''
          \},
          \{
            ``Step'': ``Second step of the problem.'',
            ``Step\_no'': ``2'',
            ``Step\_score'': ``+''
          \}
        ],
        ``Problem\_completeness\_score'': ``+'',
      \}
    ],
    ``Note\_completeness\_score'': ``+''
    \} \newline

    Here is an example: \{Example Note\}
    \newline
    \newline
    Desired output: \{Example Json\}
    \newline
    \newline
    Here is the clinical note for your task: \{Note\}\\

    \bottomrule

    \end{tabular}
    \caption{\textbf{Prompts to Gemini Pro 1.5 to transform clinical notes into JSON format.}}
    \label{tab:prompt-json}
\end{table*}

\renewcommand{\arraystretch}{1.2}
\begin{table*}[!ht]
    \small
    \captionsetup{labelfont=bf}
    \renewcommand{\tablename}{Appendix Table}
    \begin{tabular}{p{2.2cm} p{13cm}}
      
    \toprule
    \textbf{Category} & \textbf{Prompt} \\ 
    \midrule
    \textit{Prompt Template} & You are provided with a doctor-patient conversation and its corresponding clinical note in JSON format. Your task is to introduce 10 errors into the clinical note, following the instructions below.
    \newline
    \newline
    Instructions: \newline
    \{Error Type Instruction\} \newline
    - Number of Errors: Introduce 10 errors from the list above at the ``Problem'' or ``Step'' level. \newline
    - These errors can be introduced at the ``Step'' field or ``Problem'' field. \newline
    - Do not change other fields such as ``Problem\_no,'' ``Problem\_score,'' ``Step\_no,'' ``Step\_score'' or ``Problem\_completeness\_score.'' \newline
    - Recording Changes: For each change, only include the following information: \newline
        - ``Error\_type'': The type of error introduced. \newline
        - ``Problem\_no'': The number of the affected problem. \newline
        - ``Step\_no'': The number of the affected step. If change is at the problem level, output null. \newline
        - ``Error\_level'': With a value of ``Problem'' or ``Step.'' \newline
        - ``Detailed\_error'': A description of the error introduced. \newline
        - ``New\_content'': The new ``Problem'' or ``Step'' content after modification. \newline
        - ``Original\_content'': The original ``Problem'' or ``Step'' content before modification. \newline
    - Output JSON format with an ``Errors'' item only, including all information below. Do not include the original JSON file.\newline
    \newline    
    Here is the conversation: \{dialogue\} \newline
    \newline
    Here is the clinical note for your task: \{problems\} \\

    \midrule

    \textit{Factual Inaccuracy} & Error type is ``Factual Inaccuracy'': Introduce detailed factual errors related to the information or topics discussed in the conversation but not supported by it. Examples include changing ``left'' to ``right,'' altering medication names, or modifying the follow-up timeframe from ``1 month'' to ``6 months.'' \\

    \midrule

    \textit{Hallucination} & Error type is ``Hallucination'': Add completely unrelated subject entities that were not discussed in the conversation. This may include fabricated content related to symptoms, diagnostics, treatments, or other aspects. The new information should be major and entirely made up, different from minor factual inaccuracies. \\

    \midrule

    \textit{Unhelpfulness} & Error type is ``Unhelpfulness'': Rewrite sentences in a vague, incomplete, or confusing manner. Remove important details, use imprecise language, and avoid specific medical terminology or clear instructions so that the note becomes unhelpful and unclear.\\

    \bottomrule

    \end{tabular}
    \caption{\textbf{Prompts to Gemini Pro 1.5 for error generation.} Each error type is generated using different prompts by incorporating its Error Type Instruction into the prompt template.}
    \label{tab:promt-error}
\end{table*}

\renewcommand{\arraystretch}{1.2}
\begin{table*}[ht]
    \small
    \captionsetup{labelfont=bf}
    \renewcommand{\tablename}{Appendix Table}

    \begin{tabular}{p{2.2cm} p{13cm}}
      
    \toprule
    \textbf{Category} & \textbf{Prompt} \\ 
    \midrule

    \textit{Paraphrase} & You are provided with a doctor-patient conversation and its corresponding clinical note in JSON format. Your task is to introduce 20 paraphrases based on the original note, following the instructions below.
    \newline
    \newline
    Instructions: \newline
    - You want to paraphrase original sentences to improve semantic diversity of the clinical note. Please make sure the new sentence faithfully represents the same information and knowledge of the original sentence, without adding any new information. Please keep the same academic style but easy to follow, as you would expect from a medical note. \newline
    - Number of Errors: Introduce 20 different paraphrases at the ``Step'' level. Do not introduce paraphrases at the ``Problem'' level. \newline
    - It is ok to introduce multiple paraphrases for the same step. Ideally, we want to introduce paraphrases at various steps. \newline
    \newline
    ...\newline
    Here is the conversation: \{dialogue\} \newline

    Here is the clinical note: \{problems\} \\

    \bottomrule

    \end{tabular}
    \caption{\textbf{Prompts to Gemini Pro 1.5 for paraphrase generation.} Sections of the prompts related to JSON formatting, similar to those in Appendix Table \ref{tab:promt-error}, are omitted for brevity.}
    \label{tab:prompt-paraphrase}
\end{table*}

\renewcommand{\arraystretch}{1.2}
\begin{table*}[ht]
    \small
    \captionsetup{labelfont=bf}
    \renewcommand{\tablename}{Appendix Table}

    \begin{tabular}{p{2.2cm} p{13cm}}
      
    \toprule
    \textbf{Category} & \textbf{Prompt} \\ 
    \midrule

    \textit{Annotate Quality} & You are provided with a synthetic doctor-patient conversation and its corresponding clinical note in JSON format. Your task is to assess the quality of the conversation and the note.
    \newline
    \newline
    Instructions: \newline
    - Context: The conversations and clinical notes are generated by a large language model. Your task is to assess the quality of these cases by focusing on whether the conversation and notes represent real-world scenarios accurately. \newline
    - Conversation Assessment: Evaluate if the conversations are realistic and mimic true clinical interactions during outpatient visits. Be alert for unrealistic conversations, such as interactions involving a newborn speaking or inappropriate dialogue with the mother of a newborn. Identify low-quality conversations that lack sufficient detail or context. \newline
    - Clinical Note Assessment: Assess the quality of the clinical note. Identify notes that may contain inaccuracies, hallucinations not supported by the conversation, or that are incoherent or below the standard expected of high-quality medical documentation. \newline
    ...\newline
    - Recording Changes: Include only the following information in your output. \newline
        - Have two root items of ``Conversation\_quality'' and ``Note\_quality''. For each, include the following items: \newline
        - ``Rational'': Provide an explanation for your quality assessment. \newline
        - ``Quality'': Choose among ``High,'' ``Medium,'' or ``Low.'' This indicates your quality assessment. \newline
        - ``Confidence'': Choose between ``High,'' ``Medium,'' or ``Low.'' This indicates your confidence in the quality assessment. \newline
    - Output JSON format according to the specified structure. Do not include the original JSON data in your output. \newline

    Here is the conversation: \{dialogue\} \newline

    Here is the clinical note: \{problems\} \\

    \bottomrule

    \end{tabular}
    \caption{\textbf{Prompts to Gemini Pro 1.5 for quality annotation.} Notably, while we tasked Gemini Pro 1.5 with generating ratings for both conversations and clinical notes, we only used its ratings for conversations in data filtering. This decision was based on manual inspection, which revealed that its ratings and reasoning for clinical notes were of lower quality and accuracy.}
    \label{tab:prompt-eval}
\end{table*}

\renewcommand{\arraystretch}{1.2}
\begin{table*}[ht]
    \small
    \captionsetup{labelfont=bf}
    \renewcommand{\tablename}{Appendix Table}

    \begin{tabular}{p{2.2cm} p{13cm}}
      
    \toprule
    \textbf{Category} & \textbf{Prompt} \\ 
    \midrule

    \textit{Preferred Note Selection} & You are given a patient-doctor conversation and several clinical notes based on the conversation. The clinical notes only cover the ``Assessment and Plan'' section of the note. Your job is to select the best note and provide reasoning. \newline

    Here's the dialogue: \newline
    \{dialogue\} \newline

    Here are the notes: \newline
    Note 1: \newline
    \{note\_1\} \newline

    Note 2: \newline
    \{note\_2\} \newline

    Note 3: \newline
    \{note\_3\} \newline

    Provide answers in JSON format with two fields: ``Rationale'' and ``Preferred Note''. Explain your reasoning in the ``Rationale'' field step-by-step. The ``Preferred Note'' should be the number of the note you select (1, 2, or 3). Make sure your output is in valid JSON format. \\

    \bottomrule

    \end{tabular}
    \caption{\textbf{Prompts to Gemini Pro 1.5 for preferred note selection.}}
    \label{tab:preferece-selection}
\end{table*}

\begin{table*}[!ht]
\small
\renewcommand{\tablename}{Appendix Table}
\captionsetup{labelfont=bf}
\begin{tabular}{p{14.5cm}}
\toprule
\textbf{Instructions} \\
\midrule
\begin{enumerate}
\item In each row, you will be given a synthetic outpatient patient-provider dialogue from Aci-bench, and two clinical notes based on the same dialogue. We will only evaluate the ``Assessment and Plan'' parts of a note. 

\item We have performed randomization of the notes and simple processing to unify the format of notes. 

\item The dialogues from Aci-bench include conversations with (a) calls to a virtual assistant, (b) unconstrained directions or discussions with a scribe, and (c) natural conversations between a doctor and patient. Most conversations occurred in the outpatient setting.

\item Since we are focusing solely on the ``Assessment and Plan,'' you may assume that all other pertinent information from the dialogue has been documented elsewhere in the note, which is not shown here. Please evaluate the ``Assessment and Plan'' as you would in a real note. For example, relevant physical exam findings may be helpful in the ``Assessment and Plan.''

\item For each row, please start by reading the dialogue and then select your preferred notes. Make your selection based on the overall quality of the note. Essentially, choose the note you would prefer to use in a real patient encounter, imagining you are adopting AI-generated clinical notes for your daily clinical work. You may consider aspects including but not limited to: \newline
(a) Accuracy: Does the information in the clinical note accurately reflect the details from the dialogue? \newline
(b) Completeness: How well does the note cover the important information from the dialogue? \newline
(c) Helpfulness: Does this note include useful information that you would expect from a real ``Assessment and Plan''? \newline
However, remember that, ultimately, the selection should reflect your personal preference as a physician.

\item It’s perfectly acceptable to select a tie if you feel that two notes are equally good (or equally poor).

\item Please enter brief comments about each note to help us understand the rationale behind your selection. This is appreciated but not mandatory.
\end{enumerate} \\
\bottomrule
\end{tabular}
\caption{\textbf{Instructions to physicians for note preference selection.}}
\label{tab:reviewer-instruction}
\end{table*}

\renewcommand{\arraystretch}{1.2}
\begin{table*}[!ht]
    \small
    \captionsetup{labelfont=bf}
    \renewcommand{\tablename}{Appendix Table}
    \begin{tabular}{p{2.2cm} p{13cm}}
      
    \toprule
    \textbf{Category} & \textbf{Example} \\ 
    \midrule

    \textit{Top-Scoring Sample} & ASSESSMENT AND PLAN: \newline 
    1. Allergic Asthma \newline 
    Assessment: The patient has a history of allergic asthma, recently diagnosed by his primary care physician. He experiences episodic shortness of breath, eye watering, and occasional diarrhea after heavy drinking. Physical exam shows faint expiratory wheezing bilaterally in all lung fields. Chest X-ray and pulmonary function tests were normal. \newline 
    Plan: \newline 
    - Continue Albuterol Inhaler as needed. \newline 
    - Prescribe Singulair 10mg once daily. \newline 
    - Start allergy testing (skin testing) and try to identify triggers. If unsuccessful, will need further testing in the blood and possibly immunotherapy. \newline 
    Follow-up instructions: \newline 
    - Schedule follow-up appointment in one week to review results of skin testing and plan for further treatment. \\ 
    
    \midrule
    
    \textit{Negative Sample} & 
    \textbf{Problem\_1:} Newly diagnosed allergic asthma \newline 
    \textbf{Step\_1: Assessment:} Patient reports symptoms of shortness of breath, fatigue, and dry mouth after exposure to his cat. \newline 
    \textbf{Step\_score:} 0.0026 \newline 
    \textbf{Error:} Inaccuracy regarding symptoms after exposure to cat. \newline 
    ... \newline 
    \textbf{Step\_4:} Differential diagnosis includes COPD, heart failure, pneumonia, bronchitis, PFT abnormalities, vocal cord paralysis, GERD, gastroparesis, esophageal spasm, peptic ulcer disease, irritable bowel syndrome, celiac disease, lactose intolerance, Crohn's disease, ulcerative colitis, colon cancer, hemorrhoids, anal fissures, fistula, abscess, vasculitis, and sarcoidosis. \newline 
    \textbf{Step\_score:} 0.0086 \newline 
    \textbf{Error:} Hallucination about differential diagnosis. \newline 

    \textbf{Problem\_2:} Lip edema \newline 
    \textbf{Problem\_score:} 0.0420 \newline 
    \textbf{Error:} Hallucination about lip edema. \newline 
    \textbf{Step\_1: Assessment:} Patient has lip swelling and diarrhea following alcohol consumption. \newline 
    \textbf{Step\_score:} 0.2017 \newline 
    \textbf{Error:} In the dialogue the patient answered no to lip edema. \newline 
    ... \newline 

    \textbf{Problem\_3:} Nausea and vomiting after drinking \newline 
    \textbf{Problem\_score:} 0.0473 \newline 
    \textbf{Error:} Hallucination about these symptoms. In the dialogue the patient answered no to nausea and vomiting. \\ 

    \bottomrule

    \end{tabular}
    \caption{\textbf{Example 1 of clinical notes with PRM Scores.}}
    \label{tab:prm-example1}
\end{table*}

\renewcommand{\arraystretch}{1.2}
\begin{table*}[!ht]
    \small
    \captionsetup{labelfont=bf}
    \renewcommand{\tablename}{Appendix Table}
    \begin{tabular}{p{2.2cm} p{13cm}}
      
    \toprule
    \textbf{Category} & \textbf{Example} \\ 
    \midrule

    \textit{Top-Scoring Sample} & 1. ASSESSMENT AND PLAN: \newline 
    Hepatitis C \newline 
    Assessment: The patient's HCV Ab test was positive, and her liver panel showed an elevated AST (39), but her ALB, Tbili, and Total Bilirubin were all within normal limits. This suggests that she has chronic hepatitis C. \newline 
    Plan: \newline 
    - Confirm the diagnosis with additional blood work, including HCV RNA and HCV Genotype. \newline 
    - Determine the severity of liver disease with a special ultrasound. \newline 
    - Once the diagnosis and severity are confirmed, discuss treatment options with the patient. \newline
    \newline
    2. Fatigue \newline 
    Assessment: The patient reports feeling tired at the end of the day after working. She denies any other symptoms. \newline 
    Plan: \newline 
    - Recommend the patient to prioritize rest and relaxation. \newline 
    - Encourage the patient to maintain a healthy sleep schedule. \newline 
    - Consider referral to a sleep specialist if fatigue persists despite lifestyle modifications. \newline
    \newline
    Follow-up instructions: \newline 
    - Return to clinic in 3 weeks for follow-up. \\ 

    \midrule

    \textit{Negative Sample} & 
    \textbf{Problem\_1:} Hepatitis C virus (HCV) infection \newline 
    ... \newline 
    \hspace{1em} \textbf{Step\_7:} Plan: - Obtain follow-up blood tests including HCV RNA, HCV genotype, FIBROSIS OF THE LIVER, LFTS, TSH, UA, BMP. \newline 
    \hspace{1em} \textbf{Step\_score:} 0.0081 \newline 
    \hspace{1em} \textbf{Error:} Inaccuracy about tests ordered. \newline
    \hspace{1em} \textbf{Problem\_Completeness\_score:} 0.0159. \newline 
\hspace{1em} \textbf{Error:} Failed to include liver ultrasound. \newline 

    \textbf{Problem\_2:} Fatigue and mild anemia \newline 
    \hspace{1em} \textbf{Problem\_score:} 0.0110 \newline 
    \hspace{1em} \textbf{Error:} Hallucination about anemia. \newline 
    ... \newline 
    \hspace{1em} \textbf{Step\_4:} She denies any recent change in bowel movements, urination, fevers, chills, weight loss, joint pain, night sweats, shortness of breath. \newline 
    \hspace{1em} \textbf{Step\_score:} 0.2225 \newline 
    \hspace{1em} \textbf{Error:} These symptoms are not discussed in the dialogue. \newline 
    ... \newline 
    \hspace{1em} \textbf{Step\_6:} Plan: - Recommend restful activities, stress reduction techniques, counseling on healthy eating habits, exercise. \newline 
    \hspace{1em} \textbf{Step\_score:} 0.0293 \newline 
    \hspace{1em} \textbf{Error:} These plans are not discussed in the dialogue. \\ 

    \bottomrule

    \end{tabular}
    \caption{\textbf{Example 2 of clinical notes with PRM Scores}}
    \label{tab:prm-example2}
\end{table*}

\end{document}